# The syntax-semantics interface in a child's path: A study of 3- to 11-year-olds' elicited production of Mandarin recursive relative clauses

Caimei Yang[1*], Qihang Yang [1], Xingzhi Su[2], Chenxi Fu[1], Xiaoyi Wang[1], Ziman Zhuang[3], Zaijiang Man[4*]

[1] *School of Foreign Languages, Soochow University, No. 333 Ganjiang East Road, Gusu District, Suzhou, Jiangsu, China, 215006. E-mails: cmyang@suda.edu.cn (Caimei Yang);qhyang2022@stu.suda.edu.cn (Qihang Yang); 20234204028@stu.suda.edu.cn (Chenxi Fu); 20234204033@stu.suda.edu.cn (Xiaoyi Wang)*

[2] *School of Psychology, Shandong Normal University, No. 1 Daxue Road, Changqing District, Jinan, Shandong, China, 250358. E-mail: suxingzhi97@163.com (Xingzhi Su)*

[3]*Ulink High School of Suzhou Industrial Park, No.377 Linquan Street, Suzhou Industrial Park Suzhou, Jiangsu, China, 215028. E-mails: Ziman. Zhuang@sz-alevel.com (Ziman Zhuang)*

[4] *School of Linguistic Sciences and Arts, Jiangsu Normal University, No. 57 Heping Road, Yunlong District, Xuzhou, Jiangsu, China, 221009. E-mail: zjman1973@163.com (Zaijiang Man)*

*Corresponding authors:

1. Zaijiang Man;
   telephone number: 8613852003138;
   email: zjman1973@163.com;
   Postal Address:
   School of Linguistic Sciences and Arts, Jiangsu Normal University, No. 57 Heping Road, Yunlong District, Xuzhou, Jiangsu China, 221009

2. Caimei Yang;
   telephone number: 8615716203062;
   email: cmyang@suda.edu.cn;
   postal address: School of Foreign Languages, Soochow University, No. 333 Ganjiang East Road, Gusu District, Suzhou, Jiangsu, China, 215006

**Declarations of interest:** none

**Funding acknowledgement:**

This work was supported by The National Social Science Fund of China [grant numbers: 23BYY170]

**Abstract:**

**Purpose:** This study investigates the production of complex recursive relative clauses (RRCs) in Mandarin-speaking children aged 3 to 11, focusing on the syntax-semantics relationship.

**Method:** A 4 (Syntactic Types) × 2 (Semantic Conditions) × 10 (Age) design was used. The syntactic types included: (1) subject-gapped RC in object-gapped RC, (2) object-gapped RC in object-gapped RC, (3) object-gapped RC in subject-gapped RC, and (4) subject-gapped RC in subject-gapped RC. Each type was tested under irreversible internal semantics (IIS) and reversible internal semantics (RIS). For example, "the balloon that [the girl that _ eats the banana] holds _" (IIS), and "the monkey that [the dog that _ bites the pig] hits _" (RIS). Participants received speech-visual stimuli with irreversible external semantics (IES).

**Results:** The findings indicate that children produced the four types of RRCs in the IIS-IES condition one to three years earlier than their counterparts in the RIS-IES condition. This suggests a developmental progression in the acquisition of RRCs influenced by the semantic properties of the stimuli.

**Conclusions:** The results support a two-stage developmental path in the acquisition of RRCs. Initially, the language acquisition device processes the interface between syntax and IIS, eventually advancing to handle the interface between syntax and IES. These findings align with the syntax-semantics interface theory and highlight the importance of semantic context in syntactic development.

# 1. Introduction

### *1.1 Two apparently conflicting claims in the generative tradition*

In the generative tradition, there have been two apparently conflicting views over the syntax-semantics relationship in a child's path: (a) children's development of syntax is independent of any semantic requirements because syntax is autonomous and genetically endowed, which matures quickly on the condition that it is triggered by some experience-driven input which determines whether the child acquires, for example, English as opposed to German or Chinese; and (b) children's development of syntax is dependent on semantics since the accuracies of the production or interpretation of syntactically-identical-but-semantically-different expressions have been found significantly different by many studies in the literature. These two views are elaborated in the following two subsections.

#### *1.1.1 Syntax independent of semantics, and its evidence*

In theoretical linguistics, Chomsky (1957) puts forward the original doctrine of "autonomy of syntax", which is attested with the following well-known observation: Faced with an expression like *colorless green ideas sleep furiously*, people immediately know that it makes no sense in semantics but is grammatically/syntactically correct. This original doctrine has been summarized as follows: Syntax, a genetically endowed computational mechanism that yields hierarchically structured expressions, is an autonomous component of language where grammaticality exists independent of semantics. In Zhu et al.'s (2022:1) words, "a theory associated with Chomsky and others holds that syntax is a mind-internal, universal structure independent of semantics".

In the field of child language acquisition under the framework of generative grammar, the above original doctrine has led to the hypothesis that child acquisition of syntax might be independent of any semantic requirements, which is in principle compatible with the nativist approach to child language acquisition found in Chomsky’s (1957, 1981) early works, stating that children’s brains innately contain a language acquisition device which holds Universal Grammar (i.e., a biologically endowed computational mechanism that yields hierarchically structured expressions), and that this language acquisition device is independent of semantics. In this nativist tradition, since children have innately endowed knowledge of the building mechanisms of phrase structure, the work that remains for them to do is to fix the parameters of their native languages from primary linguistic data in a very short period after their birth; thus, Wexler (1998) formulates the Very Early Parameter Setting hypothesis, stating that parameters are quickly set at the very early age of a child. To support that, in line with classical studies involving the acquisition of artificial or novel expressions independent of semantics, Zhu et al. (2022) recently does experiments on materials involving artificial expressions independent of semantics to prove that infants from a very early age, say, 17 months at the latest, have abstract syntactic knowledge of their target language, e.g., the SVO structure knowledge. The Very Early Parameter Setting hypothesis aside, the nativist approach to child acquisition of syntax is in principle in line with the syntactic-bootstrapping hypothesis in Landau and Gleitman (1985), Naigles (1990), among many others, holding that innate syntax bootstraps the acquisition of the meaning of infinite novel words or expressions. For example, children use their knowledge about the syntactic category and the syntactic structure to guide the learning of the novel verb.

### *1.1.2 Syntax dependent on semantics, and its evidence*

However, the above original doctrine of "autonomy of syntax" in theoretical linguistics was recently developed into a "syntax-semantics interface" theory as stated below. The genetically endowed syntactic computational system (also called "narrow syntax") generates a potentially infinite array of hierarchically structured representations, and maps them into the corresponding semantic representations in the semantic system. If a syntactic representation can be successfully mapped into a semantic representation, the semantic interface accepts it as well-formed, otherwise, the semantic interface rejects it as malformed (Chomsky et al., 1995, 2005, 2023; Hauser et al., 2002; Berwick et al., 2013; Huybregts et al., 2016; Bolhuis et al., 2014). In a word, although still autonomous in its inner mechanism, the syntactic system must be interfaced/ compatible with the semantic system in order to generate well-formed expressions. Under this theory, both the syntactic computation and the theta-role-related semantics are universal and genetically endowed (Chomsky 2023).

This syntax-semantics interface theory is in line with many (empirical) studies in the field of child language acquisition under the framework of generative grammar; such studies usually concern natural language structures instead of artificial or novel structures. Among these studies, Roeper (1978, 2014, 2020) proposes that the syntactic faculty—which provides only an abstract framework, an idealization that does not suffice to determine the acquisition of syntax—constructs a natural syntax only in conjunction/compatibility with the semantic faculty. His supporting evidence is that children of a certain age succeed in interpreting the "semantically irreversible passive" in (1) below but fail to interpret the "semantically reversible passive" in (2); that is to say, children's interpretation accuracy between (1) and (2) is different, with the former higher than the latter (Roeper 1978)[1]. This is because, in (1), the milk, the inanimate patient, can be drunk by John, the animate agent, but not vice versa, and thus the milk and John are “semantically irreversible”, which is compatible with the irreversible nature of the syntax of English passives. However, in (2), the person namely John, can be hit by the other person, namely Bill, and vice versa, and thus John and Bill are “semantically reversible”, which is incompatible with the irreversible nature of the syntax of English passives.

(1) The milk was drunk by John.

(2) John was hit by Bill.

Roeper’s above finding of different interpretation accuracies of two passives which are syntactically identical but with a “semantic reversibility-irreversibility” contrast, has been supported by many studies using a variety of methodologies and analyses (e.g., Bever, 1970; de Villiers & de Villiers, 1973; Bridges, 1980; Stromswold, 2006). Most of these studies emphasize the inanimateness of the subject of the passive in (1) above and the animateness of the subject of the passive in (2) above, which are closely related to the “semantic irreversibility-reversibility” contrast discussed in the current paper. In a similar line, many studies (Brandt et al. 2009; Arnon 2010; Bentea et al. 2016; Kidd et al. 2007; Macdonald et al. 2020; Boudewyn et al. 2019; Nelson & Vihman 2018; Scholl & Tremoulet 2000; Vogels et al. 2013) demonstrate that single relative clauses as in (3) below are interpreted more accurately by children than their counterparts as in (4), also illustrating the “semantic irreversibility-reversibility” contrast typically resulted from the inanimateness of the head noun in (3) and the animateness of the head noun in (4).

[1]The contrast in (1) and (2) is thought to be semantic by Stromswold (2006), who names them as the “semantically irreversible passive” and the “semantically reversible passive” respectively. However, it is thought to be pragmatic in Roeper (1978), who terms them as the “pragmatically irreversible passive” and the “pragmatically reversible passive” respectively. The current study prefers the former.

(3) the ball that the boy kicked

(4) the girl that the boy kicked

The above idea that children’s syntactic faculty successfully constructs a natural syntax only in compatibility with the semantic faculty (that is, the acquisition of syntax is dependent on semantics) is in principle convergent with Wexler and Culicover’s (1980) “semantic-bootstrapping hypothesis” in child language acquisition in the generative tradition, stating that children acquire a syntax through exposure to sentences “paired with abstract hierarchical representations of their meaning”, which is also in line with Matthei’s (1982) proposal of an “isomorphic syntax-semantics relationship” ensuring the emergence/acquisition of the syntax.

***1.2 Three points warranting consideration, beyond previous studies***

The present paper aims at finding an account that can reconcile the above two apparently conflicting claims about the syntax-semantics relationship in a child’s path. To do that, we think the following three points warrant elaborate consideration, which are in principle untouched in the literature.

First, we need to consider the reason why both semantically irreversible and reversible expressions can be finally acquired in a child’s path since the former is expected and the latter unexpected under the recent “syntax-semantics interface/compatibility” theory mentioned above. To expound more clearly, as mentioned in Section 1.1.2, studies have shown that children’s accuracy in interpreting or producing semantically irreversible structures as in (1) and (3) is higher than that in interpreting or producing the semantically reversible counterparts as in (2) and (4), respectively. This is because the syntactic irreversibility of such expressions as in (1) and (3) is compatible with their semantic irreversibility, while the syntactic irreversibility of such expressions as in (2) and (4) is incompatible with their semantic reversibility. However, these previous studies fail to explain why the syntactic-irreversibility-but-semantic-reversibility incompatibility as in (2) and (4) can be finally accepted as well-formed among older children (as well as adults) since "syntax-semantics compatibility" is indispensable to the well-formedness of the expressions according to the "syntax-semantics interface" theory mentioned above.

Second, in order to find a way to solve the problem mentioned in the previous paragraph, we need to consider classifying "semantics" into two subtypes, internal and external semantics, inspired by Jackendoff (1995, 1996).To expound further, as mentioned in Section 1.1.1, *colorless green ideas sleep furiously* is considered by some researchers as an example which is "semantically" anomalous but syntactically well-formed. However, their "semantics" is in fact internal semantics, also called semantic competence or conceptual semantics in Jackendoff (1995, 1996); they fail to put *colorless green ideas sleep furiously* into such a specific, concentrated context as in Chao (1997) which gives the sentence sensible, external meaning, also called external or extensional semantics, which is external to language users as discussed in Jackendoff (1995, 1996). In a similar line, although many studies (Bever, 1970; Roeper, 1978; Bridges, 1980; Stromswold, 2006; Brandt et al., 2009; Arnon, 2010; Bentea et al., 2016; Kidd et al. 2007) discuss the "semantic irreversibility-reversibility" contrast as exemplified in (1)/(3) and (2)/(4), which are in fact related to internal semantics, they fail to take external semantics into elaborate consideration, which will be proved to be important for us to explain the reason why semantically irreversible expressions can be acquired earlier than their semantically reversible counterparts and, however, both semantically irreversible-reversible expressions can finally be acquired in a child's path.

Third, since the previous experimental studies only involve a certain relatively simpler syntax like the passive as in (1-2) or the single relative clause as in (3-4), we think we need to go beyond the previous studies to investigate the interplay effect of different complex syntaxes and different types of semantics (such as reversible and irreversible internal semantics, and irreversible external semantics). To be specific, the present paper aims to study the influence of different types of semantics on child acquisition of different types of complex recursive relative clauses[2], which has not been discussed in the literature.

To expound more clearly, children's acquisition of relative clauses (RCs) has been a hot topic in the field of language acquisition for a long time. Many studies (Sheldon, 1974; Tavakolian, 1981; Hsu et al., 2009; Hu et al., 2016; Macdonald et al., 2020; Arnon, 2010; Bentea et al., 2016; Brandt et al., 2009; Kidd et al., 2007; Adani et al., 2010; Adani et al., 2014; Haendler et al., 2015; Booth et al., 2000; Friedmann et al., 2009; Boudewyn et al., 2019; Nelson & Vihman, 2018; Scholl & Tremoulet, 2000; Vogels et al., 2013; Hsiao et al., 2022; Gennari & MacDonald, 2009; Hsiao & MacDonald, 2013;

---

[2] "Recursion has been used to characterize to the process of embedding a constituent inside another constituent of the same kind" (Martins & Fitch 2014: 17-19). In child acquisition studies, much literature (Bryant, 2006; Hollebrandse et al., 2008; Pérez-Leroux et al., 2012; Pérez-Leroux et al., 2018) deals with recursive sequences, such as recursive prepositional phrases (*the baby with the woman with the flowers*), recursive complements (*I think you said they gonna be warm*); however, there are only a few studies (Amaral & Leandro, 2018; Avram et al., 2020) investigating the acquisition of a certain type of recursive relative clauses (RRCs) like *the lion that is next to the bear that is next to the zebra*.

MacDonald & Christiansen, 2002) focus on the single RC as in (3-4) above or the single RC embedded in a main clause as in (5-6) below. Among these studies, some (Brandt et al., 2009; Arnon 2010; Bentea et al., 2016; Kidd et al., 2007; Macdonald et al., 2020; Boudewyn et al., 2019; Nelson & Vihman, 2018; Scholl & Tremoulet, 2000; Vogels et al., 2013) demonstrate that the single RC in the "semantically irreversible" condition as in (3) is interpreted more accurately by children than its "semantically reversible" counterpart as in (4). However, only several studies focus on child acquisition of a certain type of recursive RCs. For example, Amaral and Leandro (2018) investigate children's comprehension of Wapichana recursive RCs as in (7) (where a subject-gapped RC is embedded in another subject-gapped RC) and find that children are not adult-alike until they are 10 – 11 years old. Avram et al. (2020) find that 7-year-olds have not achieved adult-like comprehension of Romanian recursive RCs as in (8) (where a subject-gapped RC is embedded in another subject-gapped RC). However, all of the previous studies focusing on simple or recursive RCs fail to investigate the interplay effect of different types of complex recursive RCs and different types of semantics such as reversible and irreversible internal semantics, and irreversible external semantics.

(5) a. [The lion$_i$ that *e*$_i$ pushes the horse] knocks down the cow.

b. [The lion$_i$ that the horse pushes *e*$_i$] knocks down the cow.

(6) Xiaomei yikai-le [zuqiu dasui __ de] chuanghu

move-ASP football break DE window

'Xiaomei moved away the window [that the football broke __]'.

(7) (Py=aida un=ati) daunaiur tyka-pa-uraz [zyn kaiwada-pa-uraz kuwam].

2SG=show 1SG=to guy see-PROG-REL girl wear-PROG-REL hat

'(Show me) the guy that is seeing [the girl that is wearing a hat].'

(8) (Arată-mia) calul care este lângă [pisica care este lângă porc].

show-me horse-the that is near cat-the that is near pig

'Show me the horse that is next to [the cat that is next to the pig].'

### *1.3 The present study*

The current study addresses the above three points warranting investigation, mainly aiming to show to what extent varying types of semantics influence the acquisition of such complex syntaxes as recursive RCs, and finally to explain how varying types of semantics are integrated into a theory of syntax development which could reconcile the two apparently conflicting ideas discussed in Section 1.1.

The present study adopts a most-structured elicited production method to elicit 359 Mandarin-speaking 3- to 11-year-olds, divided into 9 groups, and 80 adults in the control group, to produce 8 target structures in a 4 (syntactic types) × 2 (semantic conditions) design as shown in Figure 1.

| | IIS-IES | RIS-IES |
|---|---|---|
| SORRCs | [$e_i$ chi xiangjiao de jiejie$_i$] na $e_j$ de qiqiu$_j$<br>eat banana De sister hold De balloon<br>'the balloon that [the sister that eats the banana] holds' | [$e_i$ yao zhu de gou$_i$] da $e_j$ de hou$_j$<br>bite pig De dog hit De monkey<br>'the monkey that [the dog that bites the pig] hits ' |
| OSRRCs | $e_i$ chi [jiejie diao $e_j$ de yu$_j$] de mao$_i$<br>eat sister fish De fish De cat<br>'the cat that eats [the fish that the sister fishes]' | $e_i$ da [zhu yao $e_j$ de gou$_j$] de hou$_i$<br>hit pig bite De dog De monkey<br>'the monkey that hits [the dog that the pig bites]' |
| SSRRCs | $e_i$ qian-zhe [$e_j$ dai maozi de gou$_j$]de gege$_i$<br>hold-ASP wear cap De dog De brother<br>'the brother that holds the dog that wears the cap' | $e_i$ da [$e_j$ yao zhu de gou$_j$] de hou$_i$<br>hit bite pig De dog De monkey<br>'the monkey that hits [the dog that bites the pig]' |
| OORRCs | [jiejie yang $e_i$ de yu$_i$] tu $e_j$ de paopao$_j$<br>sister raise De fish blow De bubble<br>'the bubbles that the fish that the sister raises blows' | [gou yao $e_i$ de mao$_i$] da $e_j$ de hou$_j$<br>dog bite De cat hit De monkey<br>'the monkey that [the cat that the dog bites] hits' |

***Figure 1 The eight targets in a 4 × 2 design (in the gloss, "De" is a relative marker; "CL" is shortened from "classifier", "ASP" "aspect" and "e" "empty category")***

To expound, Figure 1 shows four types of Mandarin prenominal recursive relative clauses (shortened as RRCs) which are different in syntax: (a) RRCs in which a subject-gapped RC is embedded in an object-gapped RC (shortened as SORRCs), (b) RRCs in which an object-gapped RC is embedded in a subject-gapped RC (OSRRCs), (c) RRCs in which a subject-gapped RC is embedded in another subject-gapped RC (SSRRCs), and (d) RRCs in which an object-gapped RC is embedded in another object-gapped RC (OORRCs). For example, the first target in the left column in Figure 1, i.e., *[$e_{i(subject\text{-}gapped)}$ chi xiangjiao de jiejie$_i$] na $e_{j(object\text{-}gapped)}$ de qiqiu$_j$* 'the balloon$_j$ that [the sister$_i$ that $e_{i(\text{subject-gapped})}$ eats the banana] holds $e_{j(\text{object-gapped})}$', is an SORRCs structure in which the prenominal subject-gapped RC *[$e_{i(subject\text{-}gapped)}$ chi xiangjiao de jiejie$_i$]* 'the sister$_i$ that $e_{i(\text{subject-gapped})}$ eats the banana' is embedded in the prenominal object-gapped RC *jiejie na $e_{j(object\text{-}gapped)}$ de qiqiu$_j$* 'the balloon$_j$ the sister holds $e_{j(\text{object-gapped})}$'. Moreover, each syntactic type in Figure 1 is put in two conditions which are different in internal semantics: reversible internal semantics (RIS) versus irreversible internal semantics (IIS). For example, in the SORRCs target mentioned above (i.e., *[$e_{i(subject\text{-}gapped)}$ chi xiangjiao de jiejie$_i$] na $e_{j(object\text{-}gapped)}$ de qiqiu$_j$*), the inanimate patient *xiangjiao* 'banana' and the animate agent *jiejie* 'sister' in the inner RC are irreversible in internal semantics, and so are the animate agent *jiejie* 'sister' and the inanimate patient *qiqiu* 'balloon' in the outer RC. Thus, this SORRCs target is in the IIS condition, compatible with its irreversible syntax. It is syntactically identical to the first target on the left column in Figure 1, i.e., *[$e_{i(subject\text{-}gapped)}$ yao zhu de gou$_i$] da $e_{j(object\text{-}gapped)}$ de hou$_j$* 'the monkey$_j$ that [the dog$_i$ that $e_{i(\text{subject-gapped})}$ bites the pig] hits $e_{j(\text{object-gapped})}$', which is also an SORRCs structure in syntax, and, however, in which *zhu* 'pig' and *gou* 'dog' in the inner RC are reversible in internal semantics (that is, both are animate and can be the agent of the predicate) and so are *gou* 'dog' and *hou* 'monkey' in the outer RC. Thus, the latter is a SORRCs structure in the RIS condition, incompatible with its irreversible syntax. The same is true with the other three pairs of targets in Figure 1, that is, OSRRCs in the IIS vs. RIS condition, SSRRCs in the IIS vs. RIS condition and OORRCs in the IIS vs. RIS condition.

It is noteworthy that, following the classical elicit production method which provides the participant with a speech-visual stimulus which is used to elicit the participant's production of the target, we provided our participants with the speech-visual stimulus which shows irreversible external semantics (IES) for each of the eight targets in Figure 1. For example, for the production of the OSRRCs structure in the RIS condition in Figure 1 (i.e., $e_{j(subject\text{-}gapped)}$ *da [zhu yao* $e_{i(object\text{-}gapped)}$ *de* $gou_i$*]* *de* $hou_j$ 'the $monkey_j$ that $e_{j(subject\text{-}gapped)}$ hits [the $dog_i$ that the pig bites $e_{i(object\text{-}gapped)}$]'), the participants were provided with the visual stimulus in Figure 2a, which presents the irreversible external semantics (IES) in that the visual stimulus clearly shows that it is the pig that bites the dog, not vice versa, and it is the monkey that hits the dog, not vice versa, which mismatches the reversible internal semantics (RIS). For another example, for the production of the SORRCs structure in the IIS condition in Figure 1 (i.e., *[*$e_{i(subject\text{-}gapped)}$ *chi xiangjiao de* $jiejie_i$*] na* $e_{j(object\text{-}gapped)}$ *de* $qiqiu_j$ 'the $balloon_j$ that [the $sister_i$ that $e_{i(subject\text{-}gapped)}$ eats the banana] holds $e_{j(object\text{-}gapped)}$]'), the participants were provided with the visual stimulus in Figure 2b, which presents irreversible external semantics (IES) matching irreversible internal semantics (IIS).

a. b.

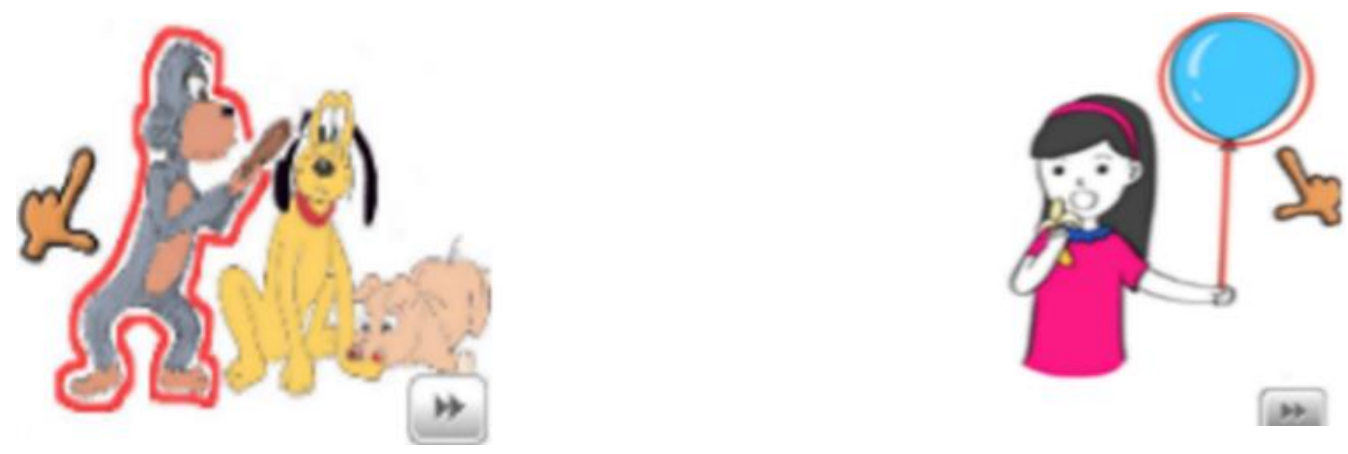

***Figure 2 (a) The visual stimulus providing IES for the OSRRCS target in the RIS condition, and (b) the visual stimulus providing IES for the SORRCs in the IIS condition***

Thus, in our 4 × 2 design, the four experimental targets in the left column in Figure 1 are all in the IIS-IES condition, in which the irreversible internal semantics (IIS) matches the irreversible external semantics (IES). For example, IIS is that a girl eats a banana (not vice versa), and the speech-visual stimulus shows IES, i.e., a girl eats a banana; in the other four targets, i.e., the SORRCs, OORRCs, OSRRCs and SSRRCs in the RIS-IES condition, the reversible internal semantics (RIS) doesn't match the IES—e.g., RIS is that a dog bites a cat and vice versa, and, however, the speech-visual stimulus shows IES, i.e., a dog bites a cat, not vice versa. The eight experimental targets are shortened as OORRCs-RIS-IES, OORRCs-IIS-IES, SORRCs-RIS-IES, SORRCs-IIS-IES, OSRRCs-RIS-IES, OSRRCs-IIS-IES, SSRRCs-RIS-IES and SSRRCs-IIS-IES in the following discussion.

The research questions in the current study are as follows:

Research question 1: Do children of a certain age succeed in producing RRCs in the IIS-IES condition but fail in its counterpart in the RIS-IES condition?

Research question 2: What are the interplay effects of different types of semantics (i.e., IIS, RIS) and different types of complex RRCs syntaxes in a child's path?

Research question 3: How are different types of semantics (i.e., IIS, RIS and IES) integrated into the acquisition of complex syntax (like RRCs) in a child's path under the framework of syntax-semantics interface theory?

## 2. Method

### *2.1 Participants*

We recruited 359 Mandarin-speaking children, aging from 3-11 years old from several kindergartens or primary schools in Tianjin and Suzhou, China, and 80 adult controls from a university in Suzhou, China. Each age group contained 43.9 participants at average, with their mean age of 3;06, 4;06, 5;05, 6;06, 7;05, 8;06, 9;05, 10;05, 11;05, 20;04 years, respectively (see Supplementary Material 2, Table 1 for more details), altogether constituting ten Age Groups. They were typically-developing and had no hearing or visual impairment.

### *2.2 Materials and design*

In this study we adopted a most-structured elicited production method as mentioned in Section 1.3, working on a 4 (Syntax) × 2 (Semantic Condition) × 10 (Age) design. The materials included 8 experimental items (aiming to elicit the 10 groups of participants to produce the eight targets in Figure 1) and 12 filler-items[3]. Each item included a speech-visual stimulus, presented with an iPad. The speech part was read in advance by a Mandarin announcer, and was presented through the speaker of the iPad in the experiment. The visual part was a flash cartoon. For example, (9) below was the speech stimulus used to elicit the participants to produce the SORRCs-IIS-IES target in Figure 1, i.e., *[$e_{i(subject\text{-}gapped)}$ chi xiangjiao de $jiejie_i$] na $e_{j(object\text{-}gapped)}$ de $qiqiu_j$* ‘the $balloon_j$ that [the $sister_i$ that $e_i$ eats the banana] holds $e_j$’. The three slides in Figures 3-5 were taken out from the visual flash cartoon matching the speech stimulus in (9), with (9a) matching Figure 3, (9b) Figure 4 and (9c) Figure 5.

(9) a. Zheli you liang-ge qiqiu,
here have two-CL balloon
‘Here are two balloons;’

b. zhe shi chi pingguo de jiejie na de nei-ge qiqiu,
this be eat apple De sister hold De that-CL balloon
‘this is the balloon that the sister that eats the apple holds;’

c. na zhe-ge ne?
then this-CL NE (NE is a sentence final participle)
‘what about this one?’

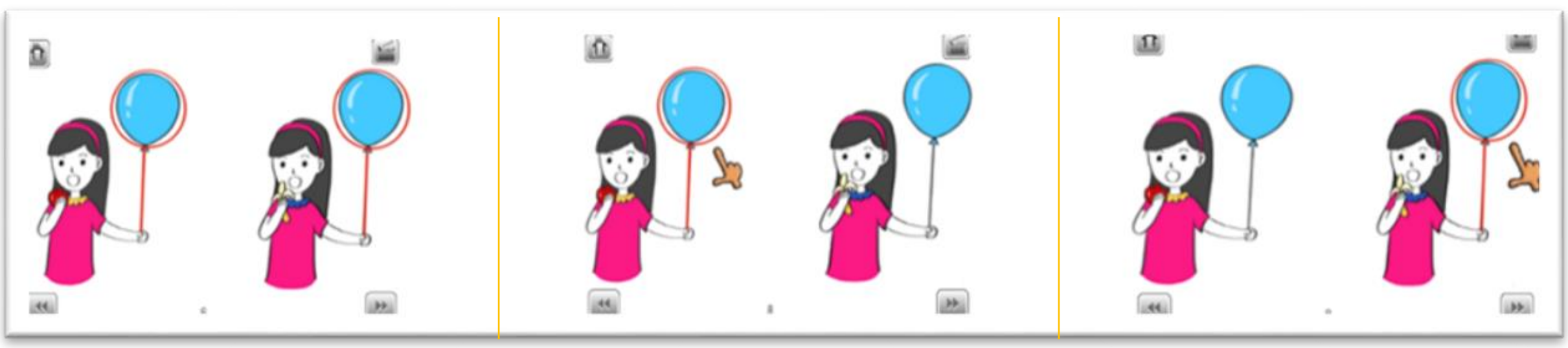

***Figures 3-5 Three slides taken from the flash cartoon matching the speech stimulus in (9a-c)***

[3]The twelve fillers involved other syntactic structures such as single RCs, RCs with passivilization, RCs with Resumptive Pronoun or RCs with clause subordination.

Similarly, (10) was the speech stimulus used to elicit the participants to produce the SORRCs-RIS-IES target in Figure 1, i.e., *[*$e_{i(subject\text{-}gapped)}$ *yao zhu de gou*$_i$*] da* $e_{j(object\text{-}gapped)}$ *de hou*$_j$ ‘the monkey$_j$ that [the dog$_i$ that $e_i$ bites the pig] hits $e_j$’. The three slides in Figures 6-8 from the visual flash cartoon matched the speech stimulus in (10).

(10) Zheli you liang-zhi houzi, zhe shi yao mao de gou da de na-zhi

here have two-CL monkey this be bite cat De dog hit De that-CL

hou, na zhe-ge ne?

monkey then this-CL N

‘Here are two monkeys; this is the monkey that the dog that bites the cat hits; what about this one?’

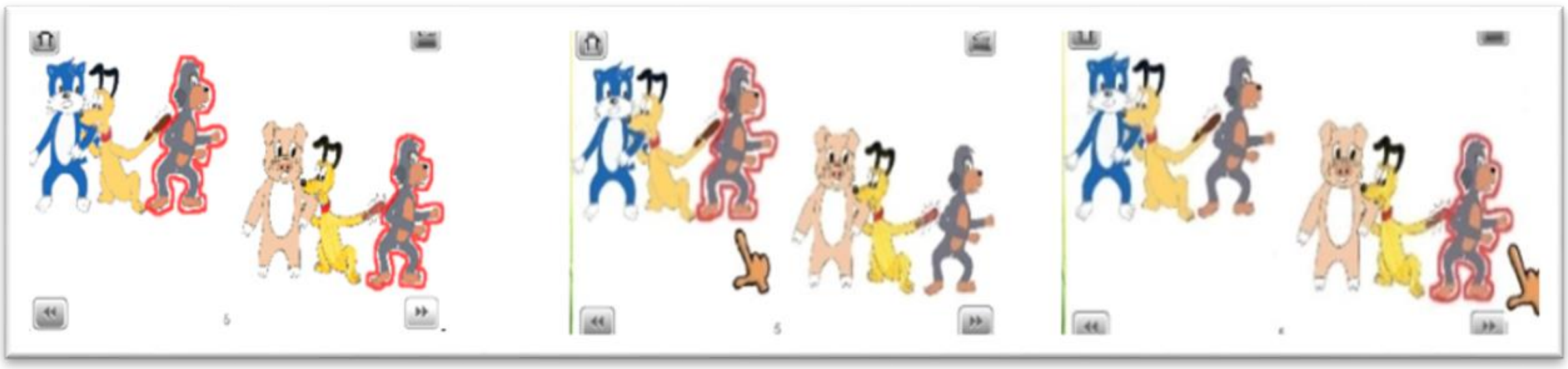

***Figures 6-8 Three slides taken from the visual stimulus paired with the speech stimulus in (10)***

It is noteworthy that, in order to create a fixed phonetic pattern which is most suitable for the elicited production of the targets and thus prevent other phonetic patterns from diverting the participants away from the targets[4], we made sure that in the speech stimuli, the SORRCs-IIS-IES prime in (9) (i.e., *[*$e_{i(subject\text{-}gapped)}$ *chi pingguo de jiejie*$_i$*] na* $e_{j(object\text{-}gapped)}$ *de qiqiu*$_j$) and the SORRCs-

[4]The lexical array in the SORRCs-IIS-IES prime *chi pingguo de jiejie na de qiqiu* ‘Lit.: eat apple De sister hold De balloon’ can be analyzed as four different syntaxes, mapping different phonetic patterns; the same is true with the lexical array in the SORRCs-RIS-IES prime *yao mao de gou da de houzi* ‘Lit.: bite cat De dog hit De monkey’ (see Figure 12a-d in Section 4.2 for details). Only one of them is the RRCs syntax used to elicit the RRCs target. Thus, the phonetic pattern of our speech prime must map the RRCs syntax.

RIS-IES prime in (10) (i.e., *[$e_{i(subject\text{-}gapped)}$ yao mao de $gou_i$] da $e_{j(object\text{-}gapped)}$ de $hou_j$*) are of a similar pattern of intonation, voice length and voice pause, as shown in Figures 9-10.

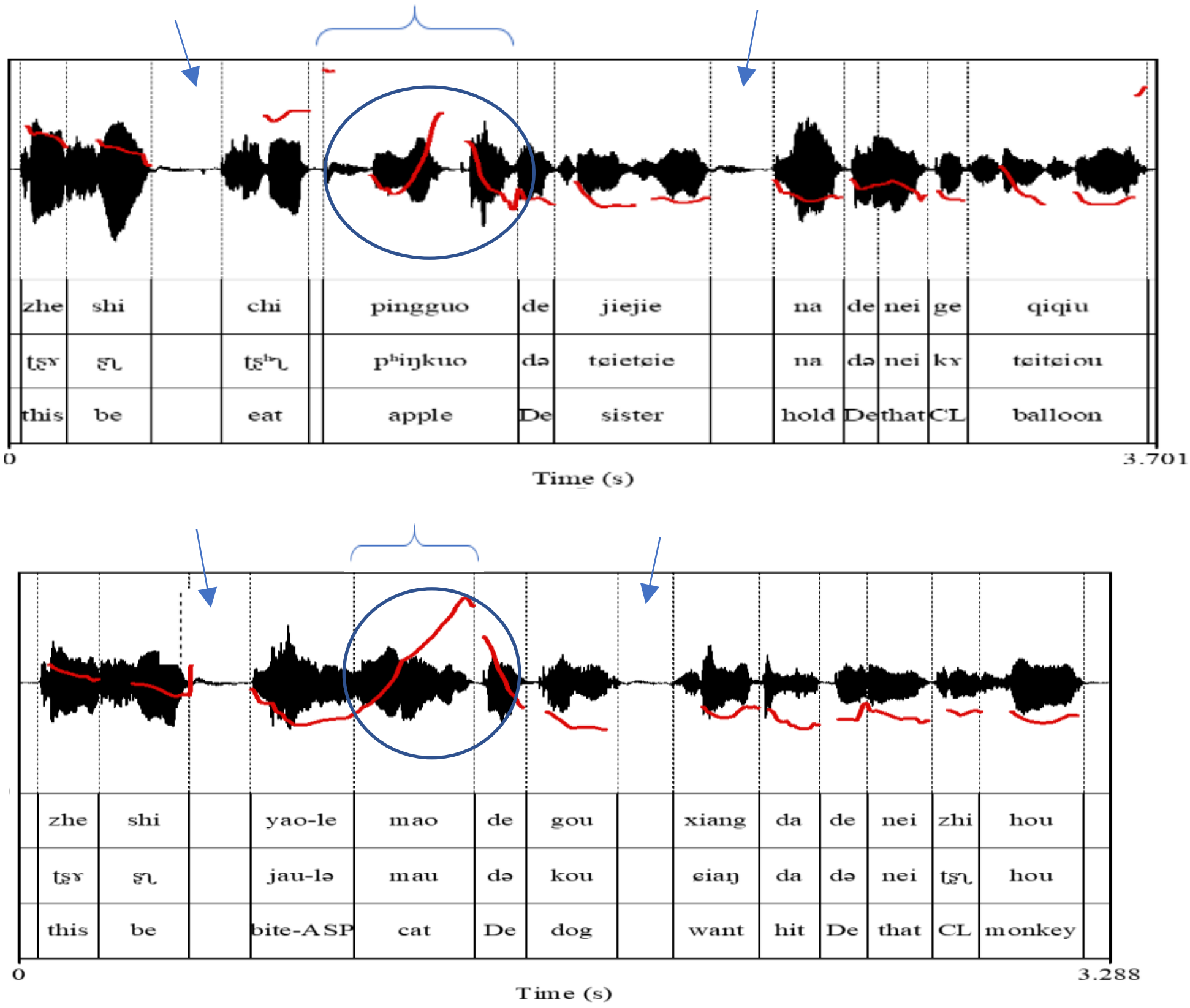

***Figure 9-10 The phonetic pattern of the prime for the SORRCs-IIS-IES target and that for the SORRCs-RIS-IES target (the curves show the pitch-related intonation; the first tier of annotation is in Chinese Pinyin, the second is in IPA and the third is the gloss)***

To be more specific, as marked by the two circles in Figures 9-10, standing against the other words' flatter background, the object *pingguo* 'apple' in the inner RC of (9) and the object *mao* 'cat' in the inner RC of (10) have a similar, highlighting intonation pattern, symbolic of a sense of detachment as well of importance and depth, since they are in the inner RC and the only words to be replaced in their corresponding target answers; for the same reason, they have the longest pronunciation, standing against the other words in their corresponding speech stimulus, as the curly brackets show. As for the voice pause pattern, there are two pauses before and after the inner RC *chi* *pingguo de jieje* ‘the sister who eats the apple’, and there are also two pauses before and after the inner RC *yao-le mao de gou* ‘the dog who bites the cat’, as the arrows show. Thus, the voice length pattern and the voice pause pattern of the pair of primes are also similar.

The other three pairs of experimental items are presented in Supplementary Material 1. Each pair is designed in a similar way to the above pair.

Taken together, our most-structured elicited production method [similar to Ambridge’s (2010) method, which elicits a response involving a simple “substitution” strategy] is also a structural priming method [similar to Pickering & Ferreira’s (2008) and Rowland et al.’s (2012)), where participants hear a prime before the production and thus are more likely to produce the primed structure than other alternatives]. This method is particularly valid for eliciting rare or complicated structures that children (as well as adults) are unlikely to produce spontaneously in daily life and for tracking the development of syntactic structures across children of different ages and adults (Ambridge & Rowland, 2013). Our experimental materials (i.e., RRCs) were rather rare and complex, some of which (e.g., the SSRRCs in the RIS-IES condition) might seldom appear spontaneously in daily life [see Karlsson (2007) and Lu (1983) for their detailed opinions on SSRRCs). Thus, since we aim to track the development of complex RRCs among 3- to 11-year-olds (with adults as controls), this most-structured elicited production method is tailor-made for the present study.

### *2.3 Procedures*

The participants were tested individually in a small room. Each of them was seated in front of the iPad to receive the speech-visual stimuli. In the training session, the objects appearing in the visual stimuli and the activities involved were made known to the participants with the help of the experimenters by asking simple questions like “What animal is this?” and “What is it/he/she doing?” or presenting statements like “There is a sister in the picture” and “The sister is eating an apple”. Following this, the experimenter started the experiment by clicking “Start” on the screen of the iPad. The speech-visual stimuli continued automatically one by one with a fixed procedure. For example, the following was the procedure for the SORRCs-IIS-IES item with the speech stimulus shown in (9) and the visual stimulus shown in Figures 3-5: First, the participant listened to the recording of (9a) *zheli you liang-ge qiqiu* ‘here are two balloons’ and at the same time watched the first part of the flash cartoon (see Figure 3) where the two balloons kept glimmering for 3 seconds; then the participant listened to (9b) *zhe shi chi pingguo de jiejie na de na-ge qiqiu* ‘this is the balloon that the sister that eats the apple holds’ and at the same time watched the second part of the flash cartoon (see Figure 4) where a “hand” appeared and kept pointing at the glimmering balloon on the left for 3 seconds; then the participant listened to (9c) *na zhe-ge ne?* ‘what about this one’ and at the same time watched the last part of the flash cartoon (see Figure 5) where the “hand” moved to point at the glimmering balloon on the right and kept pointing at it for 8 seconds, during which the participants responded. Children’s responses were recorded and transcribed.

### *2.4 Data coding*

#### *2.4.1 Data coding based on target/non-target distinction*

The targeted responses (see Figure 1) were coded as “1” and the non-targeted responses “0”. It is noteworthy that in line with previous studies (Arcodia, 2017; Tang, 2008; Chen, 2012), some variants of the targets were also coded as “1”, which are identical to the targets in syntax and meaning. As an example, as for the SORRCs-IIS-IES target *[*$e_{i(\text{subject-gapped})}$ *chi xiangjiao de jiejie*$_{i}$*] na* $e_{j(\text{object-gapped})}$ *de qiqiu*$_{j}$ ‘the balloon that the sister that eats the banana holds’, the following three variants of the target can also be coded as “1”:

Variant 1: Answers such as (11) in which the relative head is covert/deleted (the strikeout on the relative head in (11) means that the relative head can be covert/ deleted) [see more discussions in Arcodia (2017), Tang (2008) and Chen (2012)];

(11)[chi xiangjiao de ~~jiejie~~] na de ~~qiqiu~~

eat banana De sister hold de balloon

‘*(the balloon) that *(the sister) that eats a banana holds’

Variant 2: Answers such as (12) where a demonstrative (D), or a demonstrative together with a classifier, functions as a relativization marker, taking place of the typical relativization marker *de* ［see more discussions in Arcodia (2017), Tang (2008) and Chen (2012) as for the use of the demonstrative or the demonstrative with the classifier as a relative marker in Mandarin RCs, especially in spoken Mandarin];

(12)[chi xiangjiao **na** jiejie] na-zhe **na(-ge)** qiqiu

eat banana D sister hold-ASP D-(CL) balloon

‘the balloon that the sister that eats a banana holds’

Variant 3: Answers such as (13) where there is a zero-marked RC (i.e., no relativization marker). (The strikeout on the relative marker *de* in (13) means that the relative marker *de* is covert/deleted, so that the RC is zero-marked.) [5]

(13)[chi xiangjiao ~~de~~ jiejie] na de qiqiu

eat banana De sister hold De balloon

'the balloon that the sister who is eating a banana is holding'

### *2.4.2 The further coding of the non-target responses*

We distinguished the following nine types of non-target responses, coded from 0 to 9:

Code-0-type: No speech response or a response of *Wo bu zhidao* 'I don't know'.

Code-1-type: A single word such as the inner head or the outer head of the RC in the target or the substitution noun in the target. For example, instead of the SORRCs-RIS-IES target in Figure 1 (i.e., *[$e_{i(subject\text{-}gapped)}$ yao zhu de gou$_i$] da $e_{j(object\text{-}gapped)}$ de hou$_j$* 'the monkey$_j$ that [the dog$_i$ that $e_i$ bites the pig]

---

[5]The occasional omission of the relativization marker *de* can be found in adults' Mandarin (Tang 2008; Arcodia 2017). For example, in the subject-gapped RCs in (a-b) and the object-gapped RC in (c), their relative marker *de* is deleted/covert.

(a) (Haiguan chahuo shi-zhi shijie shang) **tixing** **zui-da** ~~**de**~~ **wugong**
customs seize ten-CL worldon size most-big centipedes
'(The customs seized ten) centipedes which were the biggest in size (in the world.)'
https://www.sohu.com/a/701361802_121475662 (2023.7.21)

(b) **die** **yifu** ~~**de**~~ **nvhai** (ruhe hua?)
fold clothes girl how draw
'(How do we draw) a girl who folds clothes (?)'
https://jingyan.baidu.com/article/215817f7a608f85fdb14234b.html

(c) (…fumu yaoqing-le yixie ) **bu** **renshi** ~~**de**~~ **qinqi** (lai jiali zhu)
parent invite-ASP some not know relative come home live
'(…parents invited some) relatives who I don't know (to come and live in my house)'
https://zhidao.baidu.com/question/634392583201430284.html (2023.7.21)

Thus, besides (13) mentioned in the text, we also considered (d) as a zero-marked RRCs structure.

(d) [chi xiangjiao de jiejie] na-zhe ~~de~~ qiqiu
eat banana De sister hold-ASP balloon

hits $e_j$'), the inner head *gou* 'dog' or the outer head *hou* 'monkey' or the substitution noun *zhu* 'pig' was produced.

Code-2-type: A simple declarative sentence. For example, instead of the SORRCs-RIS-IES target (see the above paragraph), a simple declarative sentence *gou yao-le zhu* 'the dog bites the pig' was produced.

Code-3-type: A conjunction of two simple declarative clauses. For example, instead of the SORRCs-IIS-IES target in Figure 1 (i.e., *[$e_{i(subject\text{-}gapped)}$ chi xiangjiao de jiejie$_i$] na $e_{j(object\text{-}gapped)}$ de qiqiu$_j$* 'the balloon$_j$ that [the sister$_i$ that $e_i$ eats the banana] holds $e_j$'), coordinated expressions such as (17) were produced.

(14) Jiejie chi xiangjiao, jiejie na qiqiu.
sister eat banana sister hold balloon
'The sister eats the banana; the sister holds the balloon.'

Code-4-type: The inner RC of the target, or an apparently deviant RC with a wrong relative head. For example, instead of producing the SORRCs-IIS-IES (see above), children produced the inner RC of the target as in (18), or an apparently deviant RC as in (19) where *qiqiu* 'balloon' is not the subject of *chi* 'eat'.

(15) chi xiangjiao de jiejie
eat banana De sister
'the sister that eats the banana'

(16) chi xiangjiao de qiqiu
eat banana De balloon
'Lit.: balloon that eats a banana'

Code-5-type: The outer RC of the target. For example, instead of producing the SORRCs-IIS-IES, children produced the outer RC of the target as in (20).

(17)jiejie na de qiqiu

sister hold De balloon

'the balloon that the sister holds'

Code-6-type: Conjoined single RCs. For example, instead of the SORRCs-IIS-IES target, the coordinated expression as in (21) was produced.

(18)jiejie chi de xiangjiao, jiejie na de qiqiu

sister eat De banana, sister hold De balloon

'the banana that the sister eats, and the balloon that the sister holds'

Code-7-type: A declarative sentence where an RC is embedded in the main clause. E.g., instead of the target SORRCs-IIS-IES, expressions as in (22) were produced[6].

(19)Chi xiangjiao de jiejie na qiqiu.

eat banana De sister hold balloon

'The sister that eats the banana holds the balloon.'

Code-8-type: The RC with a conjoined/serial verb phrase. For example, instead of the SORRCs-IIS-IES target, expressions as in (23) were produced.

(20)chi xiangjiao na de qiqiu

eat banana hold De balloon

'the balloon that (the sister) holds while eating the banana'

---

[6]To note, as pointed by Hsu et al. (2009), the expression like *nvhai xihuan xiaogou* 'Lit.: girl like puppy' produced by children can be either an unmarked object-gapped RC which means 'the puppy that the girl likes', or just a simple declarative sentence which means "the girl likes the puppy". In a similarly vein, the expression in (19) could be an unmarked RRCs structure or a declarative sentence. Thus, half of such expressions as (19) are counted as declarative sentences in Section 2.4.2.

Code-9-type: Non-targeted RRCs. For example, instead of the SORRCs-RIS-IES target (i.e., *[e$_{i(subject\text{-}gapped)}$ yao zhu de gou$_i$] da e$_{j(object\text{-}gapped)}$ de hou$_j$* 'the monkey$_j$ that [the dog$_i$ that *e*$_i$ bites the pig] hits *e*$_j$'), expressions as in (24) were produced, which had a wrong relative head.

(21)yao zhu de gou da de na-zhi mao

bite pig De dog hit De that-CL cat

'the cat that the dog that bites the pig hits'

## 3. Results and analyses

### *3.1 Results of the target responses*

The descriptive data in Figure 11 (see details in Supplementary Material 2, Figure 1) shows that, (a) in general, the percentage of the target (i.e. the accuracy percentage) for each item increased with age, (b) the accuracy percentage of SORRCs-IIS-IES, OORRCs-IIS-IES, OSRRCs-IIS-IES and SSRRCs-IIS-IES was in principle higher than that of their counterparts, i.e., SORRCs-RIS-IES, OORRCs-RIS-IES, OSRRCs-RIS-IES and SS-RIS-IES, respectively; (c) There assumed a ranking in the eight items.

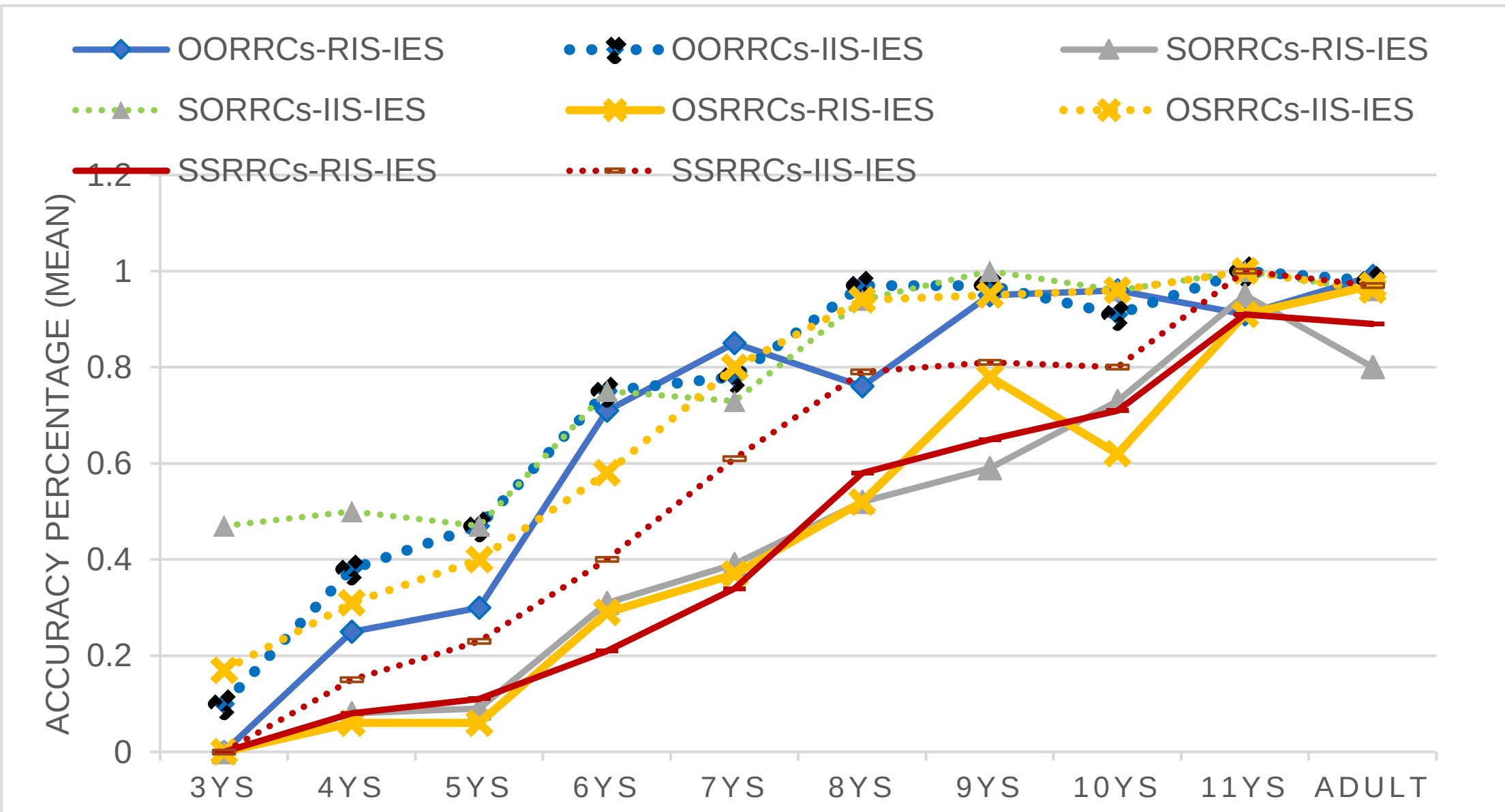

***Figure 11 Accuracy percentages of the 8 items among 10 age groups***

We conducted a generalized linear mixed effects model analysis in R (version 4.3.1, R Core Team, 2023), utilizing the blme package (Chung et al., 2013) with its default settings to implement penalization (Clark et al., 2023). The model included (Semantic) Condition, Syntax, and Age as fixed factors, and participants as random factors (Brown, 2021). Syntax was split into 4 types, Condition into 2 types, and Age into 9 groups besides the adult control group. The model found significant main effects for Age, Condition and Syntax. Follow-up pairwise tests with Bonferroni correction (see Supplementary Material 2, Table 2) for the fixed factors showed that: (a) Overall, a child could reach accuracy comparable to an adult at around 11 years of age (b = 1.26, SE = .91, $z$ = 1.39, $p$ = .163), suggesting that RRCs are rather difficult for younger children to produce; (b) the accuracy percentage of each of the RRCs in the RIS-IES condition was significantly lower than that of its counterpart in the IIS-IES condition (b = -1.98, SE=.22, z = -9.17, $p$<.001), suggesting a child of a certain age succeeds in a syntax in the IIS-IES condition but significantly fails in the same syntax in the RIS-IES condition; (c) different syntaxes correspond to significantly different response accuracy ($p$<.004), suggesting significantly varying stages/ages of syntax acquisition, with OSRRCs/SORRCs as an exception (b = -.24, SE=.22, z = -1.06, $p$=.290), which implies that these two syntaxes have a similar acquisition age.

We also summarized the results of pairwise comparisons for the significant interaction effect of Syntax-Condition and Age, with Bonferroni correction (see Supplementary Material 2, Table 3). We found that, as for the production accuracy of each structure in the IIS-IES structure, there was a significant difference between the adults and the 3/4/5/6/7-year-olds (OORRCs-IIS-IES: $ps$<.001,

OSRRCs-IIS-IES: *ps*<.001, SORRCs-IIS-IES: *ps*<.01, SSRRCs-IIS-IES: *ps*<.001), but not between the adults and the 8/9/10/11-years-olds (OORRCs-IIS-IES: *ps*⩾.498, OSRRCs-IIS-IES: *ps*≥.202, SORRCs-IIS-IES: *ps*≥.413, SSRRCs-IIS-IES: *ps*≥.056), suggesting that from age 8 on, children could produce all types of IIS-IES structures, with comparable accuracy to adults. That is to say, it needs 8 years for a child to acquire each of the four types of RRCs in the IIS-IES condition. In other words, the four syntaxes in the IIS-IES condition are acquired at the same time, i.e., age 8. However, the acquisition of the four structures in the RIS-IES condition displayed a completely different path. To be specific, we found that only 9- to 11-year-olds could produce OORRCs-RIS-IES, comparable to adults (*ps*⩾.498), only 10- to 11-year-olds could produce SORRCs-RIS-IES, comparable to adults (*ps* ⩾.126), and only 11-year-olds could produce SSRRCs-RIS-IES, comparable to adults (b =.51, SE =.76, *z* = .68, *p* =.501). As for OSRRCS-RIS-IES, there was a fluctuation: by comparison with the adults, the 9-year-olds could marginally produce it (b =1.01, SE =.55, z =-1.84, *p*=.066), while the 10-year-olds could not (b = 1.80, SE = .50, *z* = -3.06, *p* = .000), and then the 11-year-olds could (b =.27, SE =.78, z =.35, *p* =.724). Thus, as summarized in Table 1, we observed an acquisition delay of 1 years for OORRCs-RIS-IES, compared to its counterpart OORRCs-IIS-IES, a delay of 2 years for SORRCs-RIS-IES, compared with its counterpart SORRCs-IIS-IES, and a delay of 3 years for SSRRCs and OSRRCs in the RIS-IES condition, compared with their counterparts in the IIS-IES condition.

***Table 1 Acquisition ages of the eight items***

| | OORRCs | SORRCs | OSRRCs | SSRRCs |
|---|---|---|---|---|
| in the IIS-IES condition | 8ys | 8ys | 8ys | 8ys |
| in the RIS-IES condition | 9ys | 10ys | 11ys* | 11ys |

*(11ys* for the OSRRCs in the RIS-IES condition means that it might be earlier than 11 years since there is a fluctuation.)*

### *3.2 Results of the non-target responses*

Figure 12 shows that, in general, Code-5-type responses were more frequently used for each item in each age group; Code-7/8-type responses were more frequently used by the older groups (aged 7–11); Code-2-type responses were more frequently used by the younger groups (aged 3–6). Moreover, there were more Code-0/1/2/3-type responses to items in the RIS-IES condition than their counterparts in the IIS-IES condition.

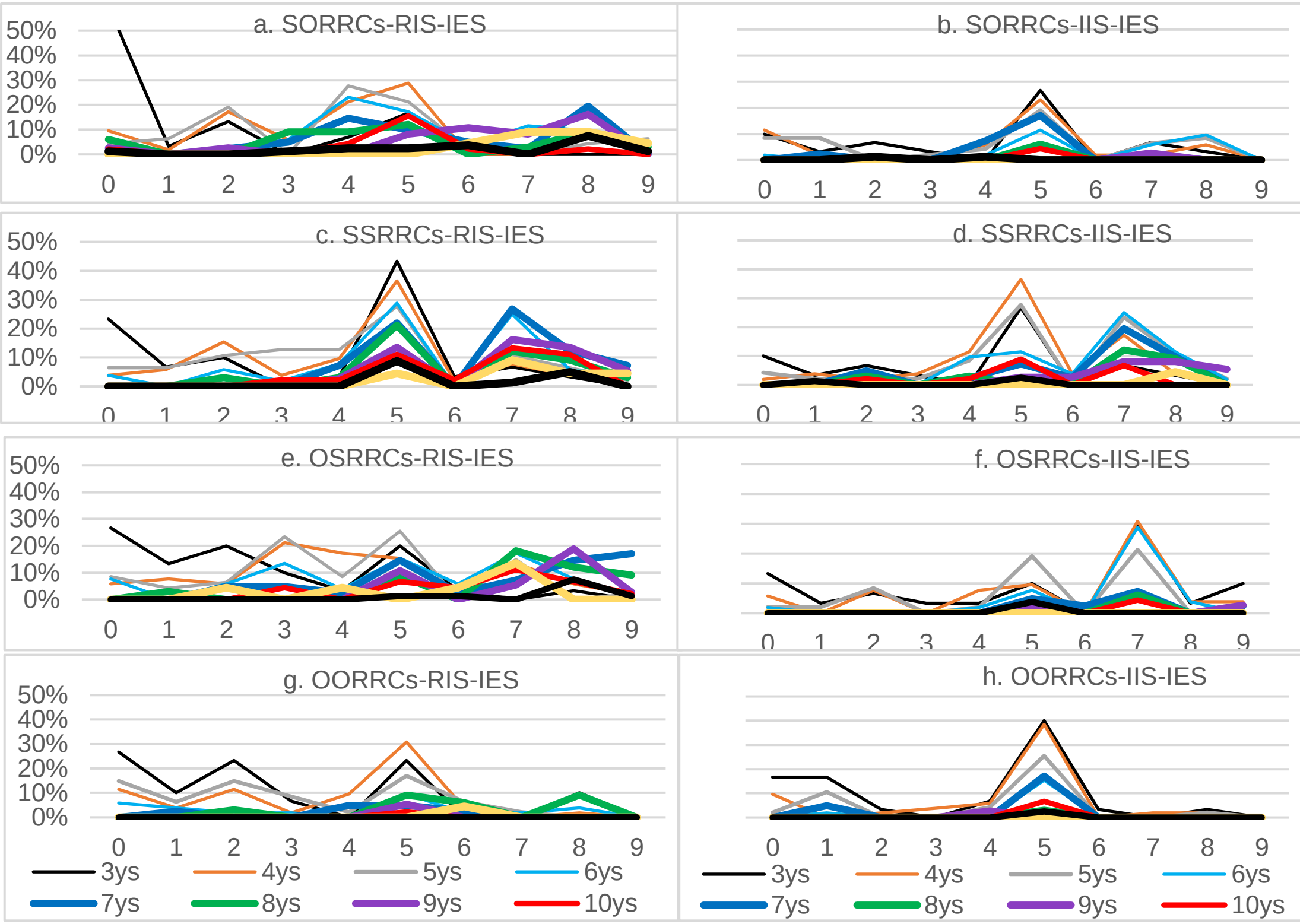

***Figures 12a–h Frequency percentage (vertical axis) of the 9 types of non-targets (horizontal axis) to the eight items (a-h) for the 10 age groups***

## 4. Discussion

### *4.1 A child's path to acquire RRCS syntaxes in the RIS-IES condition*

Our results showed that all four syntactic types of RRCs in the RIS-IES condition were acquired by children rather late; that is, at least 8-11 years was needed for children to acquire these RRCs, which is unexpected to the Wexler's (1998) Very Early Parameter Setting hypothesis. Additionally, in principle, the four RRCs were acquired at significantly different ages, i.e., at the age of 9, 10, 11*, and 11, respectively. This suggests that at the syntax level, RRCs in principle involve a type-by-type developmental path: OORRCs in the RIS condition is the first to be acquired, then SORRCs in the RIS condition, followed by OSRRCS in the RIS condition and SSRRCs in the RIS condition. This finding can be explained by several theories (Gibson, 1998, 2000; Jäger et al., 2017; McElree et al., 2003; Rizzi 2004; Yang, et al. 2022). For example, under Rizzi's (2004) Relativized Minimalism, one of the most important tasks of the language learner is to construct a syntactic algorithm to differentiate the syntactic complexity of the strings according to the distance of dependency between the gap and its co-indexed head noun and the number of noun interruption between the gap and its co-indexed noun. Figure 13 shows the distance of dependency between the gap (also called "empty category" shortened as "*e*") and its co-indexed noun, and the number of noun interruptions between the gap and its co-indexed noun. The length of the lines corresponds to the dependency distance; the arrow(s) show(s) the number of noun interruptions between the gap and its co-indexed noun. Compare the lines and arrows in the right circle, and we can see a pattern of "SSRRCs-RIS-IES>OSRRCs-RIS-IES>SORRCs-RIS-IES>OORRCs-RIS-IES" ("X > Y" means X is more complex and thus acquired later than Y), in principle compatible with our experimental results.

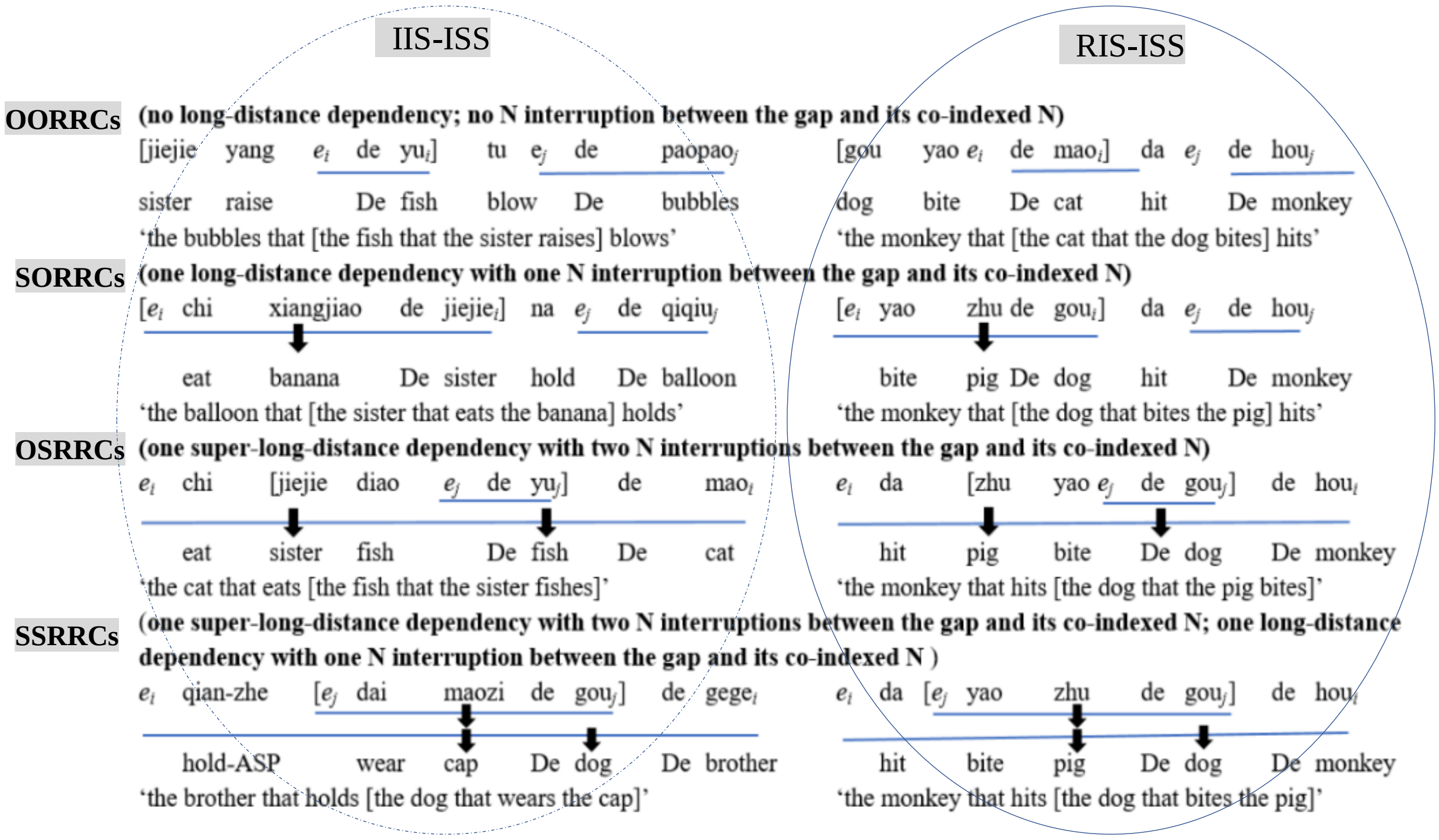

***Figure 13 The distance of dependency and the number of noun interruptions in the eight RRCs***

### *4.2 The extent to which IIS influences the acquisition of syntax in a child's path*

Compare the lines and arrows in the left dotted circle in Figure 13 above, and we should have expected a pattern of "SSRRCs-IIS-IES>OSRRCs-IIS-IES>SORRCs-IIS-IES>OORRCs-IIS-IES". However, our statistical results showed that they are acquired at the same time, and that the accuracy percentage of each of the RRCs in the IIS-IES condition was significantly higher than that of its counterpart in the RIS-IES condition, and that OORRCS, OSRRCs, OSRRCs and SORRCs in the IIS-IES condition are one to three years earlier to be acquired than their counterparts in the RIS-IES condition. These findings support the notion that irreversible internal semantics can influence and bootstrap the acquisition of syntax, especially complex syntax, in a significant way and to a significant extent. Moreover, the more complex the syntax is, the more significance the influence of irreversible semantics has on the acquisition order. In the following, we elaborate on how irreversible internal semantics influences the child acquisition of RRCs from the angle that irreversible internal semantics can disambiguate the lexical array and help create an appropriate compatibility/interface between syntax and irreversible internal semantics, taking the SORRCs-IIS-IES and SORRCs-RIS-IES items for example. The lexical array in the SORRCs-IIS-IES item *chi xiangjiao de jiejie na de qiqiu* 'Lit.: eat banana De sister hold De balloon' can be analyzed as four different syntaxes shown in Figures 14a-d (the first line). The same is true with the lexical array in SORRCs-RIS-IES *yao zhu de gou da de houzi* 'Lit.: bite pig De dog hit De monkey', also shown in Figures 14a-d (the second line).

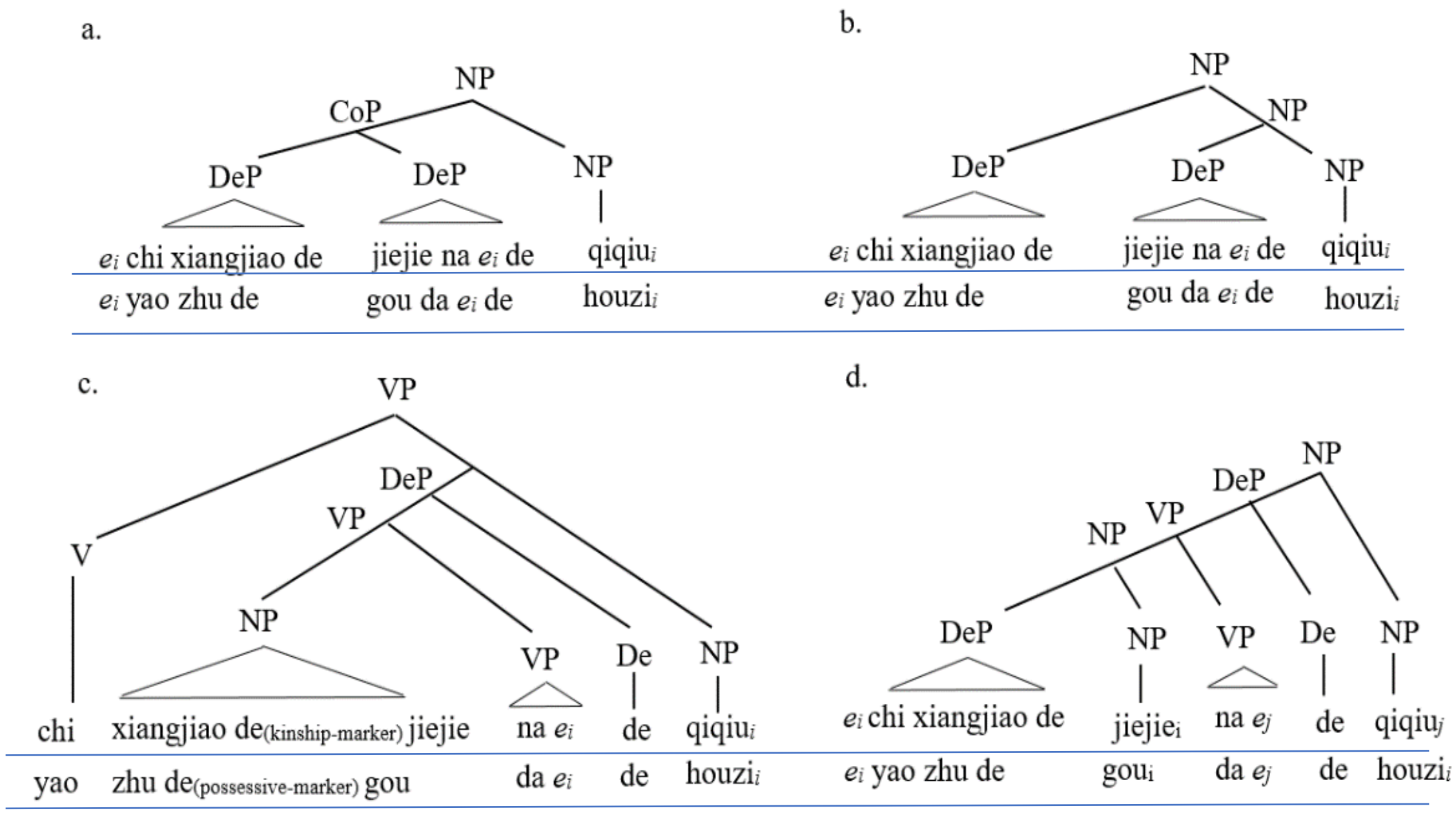

***Figures 14a–d The four syntactic structures for the lexical array of the two SORRCs items***

To expound further, the lexical arrays of the SORRCs-IIS-IES and SORRCs-RIS-IES items can be analyzed as the syntax in Figure 14a, meaning "the balloon$_i$ [which $e_i$ eats the banana and which the sister holds $e_i$]'/"the monkey$_i$ [which $e_i$ bites the pig and which the dog hits $e_i$]", or Figure 14b, meaning "[[the balloon$_i$ that the sister holds $e_i$] which $e_i$ eats the banana]'"/"[[the monkey$_i$ that the dog hits $e_i$] which $e_i$ bites the pig]", or Figure 14c, meaning "to eat [the balloon$_i$ that [ the banana's sister] holds $e_i$]"/"to bite [the monkey$_i$ that [the pig's dog] hits $e_i$])" , or Figure 14d, meaning " the balloon$_i$ that [the

sister$_j$ that $e_j$ eats the banana] holds $e_i$"/"the monkey$_i$ that [ the dog$_j$ that $e_j$ bites the pig] hits $e_i$". It is noteworthy that, for the lexical array of the SORRCs-IIS-IES item, the irreversibility of internal semantics can help directly delete the structures in Figures 14a–c, all of which obviously fail to be interpretable at the internal semantics interface (see Section 1.1.2 for the theory of syntax-semantics interface) since it is semantically impossible that a balloon eats a banana or there is a kinship between the banana and the sister. Only the structure in Figure 14d involves an interpretable/compatible interface between the irreversibility of syntax and the irreversibility of internal semantics, which is further confirmed through the irreversibility of external semantics provided by the speech-visual stimulus in the experiment, and thus the structure in Figure 14d can be interpreted/produced by children of a certain age (8-year-olds as found in the current study). However, children cannot produce the SORRCs-RIS-IES item "yao zhu de gou da de hou" 'Lit.: bite pig De dog hit De monkey' until they are 10 (as found in the current study). The reason is the following: the lexical array in the SORRCs-RIS-IES *yao zhu de gou da de hou* can also be analyzed as four syntaxes (see Figures 14a-d) and RIS makes the four syntactic analyses all interpretable at the semantic interface; thus, children of six years old are unable to choose among the four; however, when the children grow older (say, 10 years old), they can escape from the control of internal semantics and learn to create compatibility/interface between syntax and external semantics according to the speech-visual stimulus, and at this time they can choose Figure 14d which matches the external semantics.

***4.3 Three types of semantics integrated in a 2-stage development path of syntax***

Our results showed that the four types of syntaxes in the RIS-IES condition had a significant acquisition delay of 1-3 years, compared with their counterparts in the IIS-IES condition. Thus, the following 2-stage development path is hypothesized:

(a) At the first stage of the acquisition of a syntax, the acquisition device starts with seeking a compatible interface between syntax and irreversible internal semantics matching irreversible external semantics, which results in children's success in interpreting or producing expressions in the IIS-IES condition despite the failure in acquiring the counterparts in the RIS-IES condition where reversible internal semantics mismatches irreversible external semantics. At this stage, internal semantics overrides external semantics, and irreversible internal semantics essentially bootstraps syntax acquisition.

(b) At the second stage, an autonomous syntax gradually becomes stabilized, and children with this autonomous syntax can interpret or produce expressions in either the RIS or the IIS condition with the help of situations which provide children with irreversible external semantics. Therefore, at this stage, the acquisition device seeks compatibility/interface between (irreversible) syntax and irreversible external semantics, independent of children's prior internal semantics which might be incompatible with external semantics. At this stage, external semantics overrides internal semantics, and syntax-bootstrapping plays a role. Furthermore, at this stage, the acquisition of a syntax is complete. Children can even control the syntactic analyses of such apparently nonsensical expressions such as *John was drunk by the milk* by conjuring up irreversible external semantics where the milk is turned into something animate which drinks a person's blood. At this stage, much of language creativity, like poetry, depends upon the freedom to be incompatible with internal semantics.

In sum, children's acquisition device starts with seeking compatibility between (irreversible) syntax and irreversible internal semantics matching irreversible external semantics. At this first stage, expressions in the IIS-IES condition can be acquired while the counterparts in the RIS-IES condition fail to meet this compatibility and thus fail to be entered into children's syntax. The first stage of syntax acquisition needs time (sometimes a long time) to develop into a final stage where children can control a stabilized syntax and set up compatibility/interface between syntax and irreversible external semantics independent of their prior internal semantics.

***4.4 More on syntax and condition in a child's path based on non-target responses***

Our results also showed that before acquiring RRCs, younger children tended to replace them with simple structures such as a simple declarative sentence, a single RC, a conjunction of two simple declarative clauses or a conjunction of two single RCs, while older children tended to replace them with a complicated structure such as a declarative sentence with an RC embedded in it or an RC with a conjoined/serial verb phrase. In other words, the replacement was significantly influenced by age differences. We also found that the replacement was significantly influenced by the IIS/RIS condition contrast. These results suggest several theoretical implications. First, in children's developmental path, their syntax develops step by step from less to more hierarchical/complicated structures. Second, the children will not avoid more complicated structures that they have already grasped and thus the percentage of simple structures decreases with age while that of more complicated structures increases with age. Third, the triggering effect of IIS on the development of syntax also works in the production of non-target answers. Fourth, children's syntax may be more flexible than that of adults since they frequently produce the apparently deviant single RC. This can be demonstrated with several participants' response shown in (22), which is semantically unacceptable to adults. This can also be demonstrated with a participant's response as shown in (23), which is never thought of as acceptable in adults' syntax. It is noteworthy that the visual stimulus shows that *gege* 'brother' does not wear a cap and from the context in (23) we can also see that the child knows that it is the dog that wears the cap.

(22) chi xiangjiao de qiqiu (repeated from (19) in Section 2.4.2)

eat banana De balloon

'Lit.: balloon that eats a banana'

(23) A participant's response for SS-IIS-IES: (age: 4;10)

Tade gou dai maozi, suoyi ta shi dai maozi de gege

his dog wear cap so he be wear cap De brother

'Lit.: his dog wears a cap, so he is the brother (whose dog) wears a cap'

## 5. Conclusion

Our results confirm our proposal that child acquisition of syntax develops from the first to the second stage. At the first stage, the acquisition device seeks the minimal/compatible interface between (irreversible) syntax and irreversible internal semantics. However, this stage is not stable in that children can acquire a structure in the IIS-IES condition but fail to acquire its counterpart in the RIS-IES condition in which the interface between syntax and reversible internal semantics is not minimal. At the second stage, an autonomous syntax is stabilized and children at this stage can create a minimal interface between syntax and irreversible external semantics (just like adults) based on the situations they are facing or they are imagining, independent of their prior irreversible or reversible internal semantics. Moreover, most previous experimental studies on child's interpretation or production of syntactic structures design experiments involving visual stimuli which present external semantics; however, the present study shows that the participant may be too young to be influenced by the visual stimuli which provide external semantics. The present study shows that the participants can be influenced by the visual stimuli only at the second stage of acquisition.

Click here to access/download

**Supplemental Material**

Supplemental Material 1.docx

Click here to access/download

**Supplemental Material**

Supplementary Material 2.docx

Click here to access/download
**Supplemental Material**
Original data including coding.xlsx

Click here to access/download
**Supplemental Material**
data coding based on target&non-target distinction.xlsx